\documentclass[runningheads]{svjour3}
\smartqed  
\usepackage{graphicx}
\usepackage{fix-cm}
\usepackage{graphicx}
\usepackage{amsmath}
\usepackage{amssymb}
\usepackage{amsfonts}
\usepackage{latexsym}
\usepackage{subfigure}

\newcommand {\R} {\mathbb{R}}
\newcommand {\Om} {\mathcal{O}}
\newcommand {\grad} { { \boldsymbol \nabla} }

\journalname{Mathematical Programming}
\begin{document}

\title{Faster gradient descent and the efficient recovery of images}


\author{ Hui Huang \and Uri Ascher }
      
\institute{Hui Huang \at Shenzhen Institute of Advanced Technology, Chinese Academy of Sciences, China\\ 
           \email{hui.huang@siat.ac.cn}          
           \and
           Uri Ascher \at Department of Computer Science, University of British Columbia, Vancouver, Canada\\
           \email{ascher@cs.ubc.ca}
}

\date{Received: May 2009 / Accepted: December 2010}

\maketitle

\begin{abstract}
Much recent attention has been devoted to gradient descent algorithms where the
steepest descent step size is replaced by a similar one from a previous iteration
or gets updated only once every second step, thus forming a 
{\em faster gradient descent method}.
For unconstrained convex quadratic optimization these methods can converge much faster
than steepest descent. But the context of interest here is application to certain
ill-posed inverse problems, where the steepest descent method is known to have a smoothing,
regularizing effect, and where a strict optimization solution is not necessary. 

Specifically,
in this paper we examine the effect of replacing steepest descent
by a faster gradient descent algorithm
in the practical context of image deblurring and denoising tasks.
We also propose several highly efficient schemes for carrying
out these tasks independently of the step size selection,
as well as a scheme for the case where both blur and significant noise are present.

In the above context there are situations where many
steepest descent steps are required, thus building slowness into the solution procedure.
Our general conclusion regarding gradient descent methods
is that 
in such cases the faster gradient descent methods
offer substantial advantages. In other situations where no such slowness buildup
arises the steepest descent method can still be very effective.

\end{abstract}

\section{Introduction}
\label{sec:intro}

The tasks of deblurring and denoising are fundamental in image restoration and have received
a lot of attention in recent years; see, e.g., \cite{vogelbook,chsh,of,dafrmo,mallat} and references therein.
These are inverse problems, and they can each be formulated as the recovery
of a 2D surface model $m$
from observed (or given) 2D data $b$ based on the equation
\begin{eqnarray}
b = F(m) + {\epsilon}.
\label{1}
\end{eqnarray}
Here $F(m)$ is the predicted data, which is
a linear function of the sought model $m$, and
$\epsilon$ is additive noise. Both $m$ and $b$ are defined on a rectangular pixel grid,
and we assume without loss of generality that this grid is square, discretizing
the unit square $\Omega$ with $n^2$ square cells of length $h= 1/n$ each.
Thus, $m = \{m_{i,j}\}_{i,j=0}^n$.

In the sequel it is useful to consider $m$ in two other forms.
In the first of these, $m$ is reshaped into a vector in
$\R^N, \;  N = (n+1)^2$.
Doing the same to $b$ and $F$, we can write the latter as a matrix-vector multiplication,
given by
\begin{eqnarray}
F(m) = Jm,
\label{Jm}
\end{eqnarray}
where the {\em sensitivity} matrix $J = \frac{\partial F}{\partial m}$ is constant.
Note that $N$ can easily exceed $1,000,000$ in applications; however,
there are fast matrix-vector multiplication algorithms available to calculate $F$
for both deblurring \cite{vogelbook} and (trivially) denoising.

The second form is really an extension of $m$ into a piecewise smooth function $m(x,y)$
defined on $\Omega$. This allows us to talk about integral and differential terms in
$m$, as we proceed to do below, with the understanding that these are to be discretized
on the same grid to attain their true meaning.

Both inverse problems are ill-posed, the deblurring one being so even without the presence
of noise \cite{chsh}. Some regularization is therefore required.
Tikhonov-type regularization is a classical way to handle this, leading to the
optimization problem
\begin{eqnarray}
\min_m\;  T(m; \beta) \equiv \frac 12 \| Jm - b \|^2 + \beta R(m),
\label{TikhJm}
\end{eqnarray}
where $R(m)$ is the regularization operator and $\beta > 0$ is the
regularization parameter \cite{ehn1,ta}. It is important to realize
that the determination of $\beta$ is part of
the regularization process. The least-squares
norm $\|\cdot\|$ is used in the data-fitting term for simplicity, and it corresponds
to the assumption that the noise is Gaussian. The necessary
condition for optimality in \eqref{TikhJm} yields the algebraic system
\begin{eqnarray}
G(m) \equiv \nabla_m T(m; \beta) \equiv  J^T (Jm - b) + \beta R_m = 0,
\label{nece}
\end{eqnarray}
where $R_m$ is te gradient of $R$.

The choice of $R(m)$ should incorporate {\em a priori} information, such as piecewise
smoothness of the model to be recovered, which is the vehicle for noise removal.
Consider the one-parameter family of Huber switching functions, whereby $R$
is defined as the same-grid discretization of
\begin{subequations}
\begin{eqnarray}
R(m) &=&  \int_\Omega \rho (| \grad m | ), \label{regR}\\
\rho(\sigma) = \rho (\sigma; \gamma) &=& \begin{cases} \sigma, & |\sigma| \geq \gamma  \cr
                     \sigma^2/(2\gamma) + \gamma/2, & |\sigma| < \gamma
                \end{cases}  .\label{huber}
\end{eqnarray}
\label{hubereg}
\end{subequations}
The corresponding necessary conditions \eqref{nece} are the same-grid discretization of
an elliptic PDE with the leading differential term given by
\begin{eqnarray}
R_m ({m}) &=& L(m) \cdot m, \nonumber \\
L(m) &=& -\nabla \cdot \left( \frac 1{|\grad m|_\gamma}\grad \right), \quad
|\grad m|_\gamma = \max\{\gamma, |\grad m|\}.
\label{Lm}
\end{eqnarray}
Thus, for $\gamma$ large enough so that always $ \max\{\gamma, |\grad m|\} = \gamma$ the
objective function is a convex quadratic and \eqref{nece} is a linear symmetric positive definite system.
However, this choice smears out image edges. At the other extreme, $\gamma = 0$ yields
the total variation (TV) regularization \cite{clmu,rof,vogelbook,fo}. But this requires modification when
$m$ is flat to avoid blowup in $L$. For most examples reported here we employ the adaptive choice
\begin{eqnarray}
\gamma =  \frac{h}{|\Omega|} \int_\Omega | \grad m| ,
\label{gammadef}
\end{eqnarray}
proposed in \cite{ahh}, which basically sets a resolution-dependent switch, modifying TV.

For the approximate solution of the optimization problem \eqref{TikhJm}, consider combining
two iterative techniques. The first iteration applies for the case where $\gamma$ is
small enough so that $R_m$ is nonlinear in $m$, as is the case when using \eqref{gammadef}.
In this case a fixed point iteration called lagged diffusivity (or IRLS)
is employed \cite{vogelbook}, where $m^0$ is an initial guess and $m^{k+1}$ is defined as
the solution of the linear problem
\begin{eqnarray}
J^T(Jm - b)   + \beta L(m^{k}) m = 0,
\label{lagdif}
\end{eqnarray}
for $k=0,1,2,\ldots $. See \cite{ahh,chmu} for a proof of global convergence of this iteration.
In practice a rapid convergence rate is observed at first, typically slowing down only in a regime
where the iteration process would be cut off anyway.

Next, for the solution of \eqref{TikhJm} or the potentially large linear system
\eqref{lagdif} consider the iterative method of gradient descent
given by
\begin{eqnarray}
m^{k+1} = m^{k} - \tau_k \left( J^T(Jm^k - b) + \beta L(m^{k}) m^k \right) , \quad k=0,1,2,\ldots .
\label{gradesc}
\end{eqnarray}
Such a method was advocated for the denoising problem already in \cite{rof,pema}, 
and it corresponds to a forward Euler
discretization of the embedding of the elliptic PDE in a parabolic PDE with artificial time $t$,
written as
\begin{eqnarray}
\frac {\partial m}{\partial t} = -\left( J^T (Jm - b) + \beta R_m \right), \quad t \geq 0.
\label{neceart}
\end{eqnarray}
See also \cite{gigo}.


The step size $\tau_k$ in \eqref{gradesc} had traditionally been determined
for the quadratic case by exact line search, yielding the steepest descent (SD) method.
But this generally results in slow convergence, as slow as when using the best uniform step size,
unless the condition number of $J^TJ + \beta L$ is small
\cite{akaike,nosazh,asdohusv}. The convergence rate then is
far slower than that of the method of conjugate gradients (CG).
In recent years much attention has been paid to gradient descent methods 
where the steepest descent step size value from the previous 
rather than the current iteration
is used \cite{babo}, or where it gets updated only once every second iteration
\cite{rasv,fmmr}. These step selection strategies yield in practice a much faster convergence for the
gradient descent method applied to convex quadratic optimization, although they are
still slower than CG and their theoretical properties are both poorer and more mysterious \cite{asdohusv}.
Let us refer to these variants as {\em faster gradient descent} methods.
These methods automatically combine an occasionally large step that may severely violate
the forward Euler absolute stability restriction but yields rapid convergence
with small steps that restore stability and smoothness. The resulting dynamical system is chaotic \cite{doas3}. 

Faster gradient descent methods have seen practical use in the optimization context of
quadratic objective functions subject to box constraints \cite{dafl}, especially
when applied to compressed sensing and other image processing problems \cite{finowr,befr}.
For unconstrained optimization they are generally majorized by CG,
and the same holds true for their preconditioned versions.
Here, however, the situation is more special.
For one thing, the PDE \eqref{neceart} has in the case of denoising, where $J$ is the identity,
the interpretation of
modeling the physical process of anisotropic diffusion
(isotropic for sufficiently large $\gamma$). This has a smoothing effect that is desirable
for the denoising problem. CG is also a smoother \cite{hansen}, but it does not have the same physical interpretation.
Moreover, CG is more susceptible to perturbation caused by the lagged diffusivity or IRLS method,
where the quadratic problem solved varies slightly from one iteration to the next (see also \cite{doas3}).
Related and important practically, the iteration process \eqref{gradesc} need not
be applied all the way to strict convergence, because a desirable regularization effect is obtained already
after relatively few iterations.
To understand this, recall that the parameter $\beta$ in \eqref{TikhJm} is still to be determined.
Its value affects the Tikhonov filter function and relates to the amount of noise present \cite{vogelbook}.
But it is well known that an alternative to the Tikhonov filter function is the exponential filter function---see, e.g., \cite{vogelbook,care}---and the
latter is approximately obtained by what corresponds to integrating \eqref{neceart}
up to only a {\em finite time}; see \cite{ashudo,gigo} and references therein.
In fact, upon integration to finite time one may optionally set $\beta = 0$, replacing the Tikhonov
regularization altogether.
The question now is, in the current problem setting, do faster gradient descent methods perform better than
steepest descent, taking larger steps and yet maintaining the desired regularization effect at the same time?
More specifically, does the more rapid convergence provided by the large steps and the regularization
(i.e., piecewise smoothing) effect provided by the small steps automatically combine to form
a method that achieves the desired effect in much fewer steps? 


In \cite{asdohusv} we started to answer this question for the denoising problem.
Here we continue and extend that line of investigation. In Section~\ref{sec:denoising}
we complete the results presented in \cite{asdohusv} and propose a new, efficient,
explicit-implicit denoising scheme with an edge sharpening option. 

In Section~\ref{sec:deblurring}, the main section of this article, we apply
the methods described above to the much harder deblurring problem, 
and also propose a new method for the case
where there are both blur and significant noise present.

The two step size selections that are being compared for the deblurring problem
can be written as
\begin{subequations}
\label{steps}
\begin{eqnarray}
\mbox{SD:} \qquad  \tau_k &=& \frac{(G(m^k))^TG(m^k)}{(G(m^k))^T(J^TJ + \beta L(m^k))G(m^k)}, \label{sd}\\
\mbox{LSD:} \qquad   \tau_k &=& \frac{(G(m^{k-1}))^TG(m^{k-1})}{(G(m^{k-1}))^T(J^TJ + \beta L(m^{k-1}))G(m^{k-1})}
\label{lsd} .
\end{eqnarray}
These are the steepest descent (SD) and lagged steepest descent (LSD) formulas for the case 
of large $\gamma$, where 
the matrix $-L$ is just the discretized Laplacian subject to natural boundary conditions, 
and they play a similar role for the nonlinear case in the context of
lagged diffusivity.
\end{subequations}

All presented numerical examples were run on an Intel Pentium 4 CPU 3.2 GHz machine with
512MB RAM, and CPU times are reported in seconds.
Conclusions are offered in Section~\ref{sec:conclusion}.

\section{Denoising}
\label{sec:denoising}

For the denoising problem we have $F(m) = m$, so the sensitivity
matrix $J$ is the identity, and it is trivially sparse and
well-conditioned. In \cite{asdohusv} we have considered the
well-known gradient descent algorithm
\begin{subequations}
\label{fesd}
\begin{eqnarray}
m^0 &=& b,  \label{fesda}\\
m^{k+1} &=& m^k - \tau_k R_m(m^k), \quad k = 0, 1, 2, \ldots ,  \label{fesdb}
\end{eqnarray}
\end{subequations}
obtained as a special case of \eqref{gradesc} upon rescaling the artificial time by $\beta$ and then letting
$\beta \rightarrow \infty$. This gets rid of the annoying need to determine $\beta$.
The influence of the data is only through the initial conditions,
i.e., it is a pure diffusion simulation.
Correspondingly, in the step size definitions \eqref{steps}, $G$ is replaced by $R$
and $J^TJ + \beta L$ is replaced by $L$.

The results in \cite{asdohusv} clearly indicate not only the
superior performance of the parameter selection \eqref{gammadef}
over using large $\gamma$ but also the improved speed of convergence
using LSD over SD, occasionally by a significant factor. For instance, cleaning
the {\tt Cameraman} image used below in Fig.~\ref{camhy}, which is $256 \times
256$ and corrupted by $20\%$ Gaussian white noise, agreeable results
are obtained after $382$ iterations using SD as compared to only
$117$ iterations using LSD. These numbers arise upon using the same relative error norm,
defined by
\begin{eqnarray}
e^k = \|m^{k+1}-m^k\|/\|m^{k+1}\|,
\label{relaerr}
\end{eqnarray}
to stop the iteration in both cases. Unless otherwise noted 
we have used a sufficiently strict tolerance to 
ensure that the resulting images are indistinguishable.
Similar results are obtained for other
test images.

Still, these methods can often be further improved. With LSD,
occasionally the larger step sizes could produce a slightly rougher image
than desired (compare Figs.~4(f) and 4(e) in \cite{asdohusv}),
whereas SD occasionally simply takes too long. 
We may therefore wish to switch to solving the Tikhonov equations
\eqref{nece}, which here read
\begin{eqnarray}
m - b + \beta R_m = 0.
\label{necedenoise}
\end{eqnarray}
However, this brings back the question of effectively selecting the parameter $\beta$.

Fortunately, there is a fast way to determine $\beta$, at least for the
case where the noise is Gaussian.
Using \eqref{hubereg} with \eqref{gammadef}, the reconstructed model
becomes much closer to the true image than to the noisy one already after
a few LSD or even SD iterations, see \cite{asdohusv}.
The computable misfit, defined by
\begin{eqnarray}
 \eta_k = \| m^k - b \| /n ,
 \label{hui1b}
\end{eqnarray}
therefore provides a very good, cheaply obtained approximation for the noise level.
Thus, denoting the roughly denoised image by
$\bar m$, we have $\int_{\Omega}|\bar m-b|^2 \approx \eta^2$.
According to the discrepancy principle, we wish our intended reconstruction $m$ to
maintain this constraint invariant in ``time'', that is
$$\frac{d}{dt}\int_{\Omega}|m(t)-b|^2 = 0,$$
which by \eqref{neceart} yields
\begin{eqnarray*}
0 = \int_\Omega(m-b)\frac{\partial m}{\partial t} = -\left( \int_\Omega |m-b|^2 + \beta \int_\Omega (m-b)R_m \right).
\label{conbeta}
\end{eqnarray*}
This therefore leads us to determine $\beta$ as
\begin{eqnarray}
\beta = -\frac{\int_\Omega |\bar m-b|^2}{\int_\Omega (\bar m-b)R_m(\bar m)} .
\label{eibeta}
\end{eqnarray}
See also \cite[Section~11.2]{of}.

Now that we have $\beta$ (and $\bar m$ for an initial guess)
we solve the equations \eqref{necedenoise} with \eqref{hubereg}
and \eqref{gammadef},
which corresponds to an implicit artificial-time integration step,
using a combination of lagged diffusivity (IRLS) and CG with a multigrid preconditioner;
see \cite{ahh} for details.

\begin{figure}[tp]
\centering 
\subfigure[True image]
{\label{lena}\includegraphics[scale=0.35]{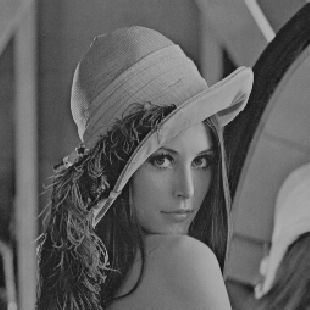}}
\subfigure[10\% noise]
{\label{lena10}\includegraphics[scale=0.35]{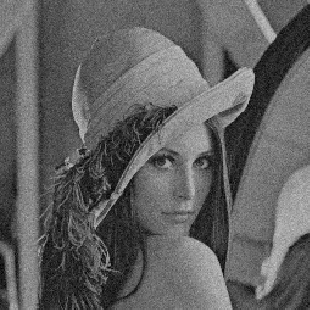}}
\subfigure[Misfit = 9.74]
{\label{lenaSDe-4}\includegraphics[scale=0.35]{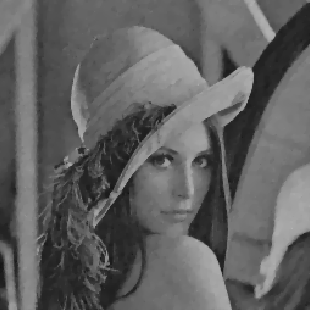}}
\centering \subfigure[Misfit = 11.17]
{\label{lenaSD}\includegraphics[scale=0.35]{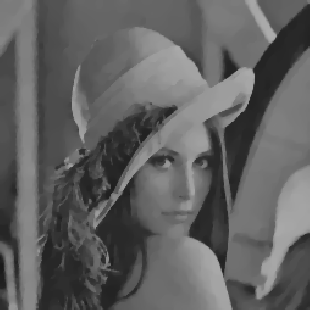}}
\subfigure[Misfit = 11.93]
{\label{lenaim}\includegraphics[scale=0.35]{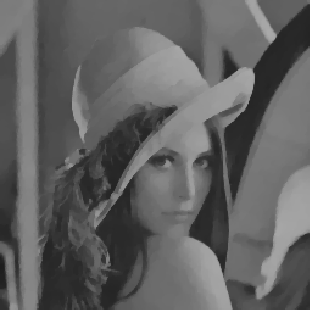}}
\subfigure[Misfit = 10.39]
{\label{lenatu}\includegraphics[scale=0.35]{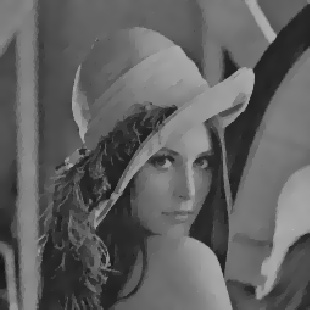}}
\caption[Comparing different denoising schemes]{Comparing different
denoising schemes on the $265 \times 256$ {\tt Lena} image: (a) true
image; (b) image corrupted by $10\%$ Guassian white noise serves as
data; (c) image denoised using SD step sizes, stopping when $e^k$ is
below $10^{-4}$, 17 iters; (d) image denoised using SD step sizes,
stopping when $e^k$ is below $10^{-5}$, 309 iters, 51.9 sec; (e)
image denoised by 3 IRLS iterations with $\beta = 0.083$ after 17 SD
steps, the total CPU time = 2.7 + 5.2 = 7.9 sec; (f) image sharpened
using 10 SD steps with Tukey function after 17 pre-denoising SD
steps with Huber function, the total CPU time = 2.7 + 2.2 = 4.9
sec.} \label{lenahy}
\end{figure}

In Figs.~\ref{lenaSD} and \ref{lenaim}, we compare the result of
this hybrid explicit-implicit scheme with that of the pure explicit
scheme \eqref{fesd}. The pre-denoised image after 17 SD steps, shown in
Fig.~\ref{lenaSDe-4}, acts as a warm start and produces a good
regularization upon setting $\beta = 0.083$ by \eqref{eibeta} for the
following implicit process. Then, after only 3 IRLS iterations, the
denoised image in Fig.~\ref{lenaim} looks already quite comparable
with---even slightly smoother than---the one in Fig.~\ref{lenaSD},
which is continually denoised using SD for a stricter relative error tolerance. The
processing time of the hybrid scheme is only a small fraction of
that required by the explicit scheme, even with the faster LSD step
size. Figs.~4(e) in \cite{asdohusv} and \ref{camim} in the present article 
tell us essentially the same story for a different example.
\vspace{0.7cm}

\noindent{\bf Sharpening the reconstructed image}
\vspace{0.3cm}

\begin{figure}[tp]
\centering \subfigure[Huber switching function]
{\label{Rhuber}\includegraphics[scale=0.53]{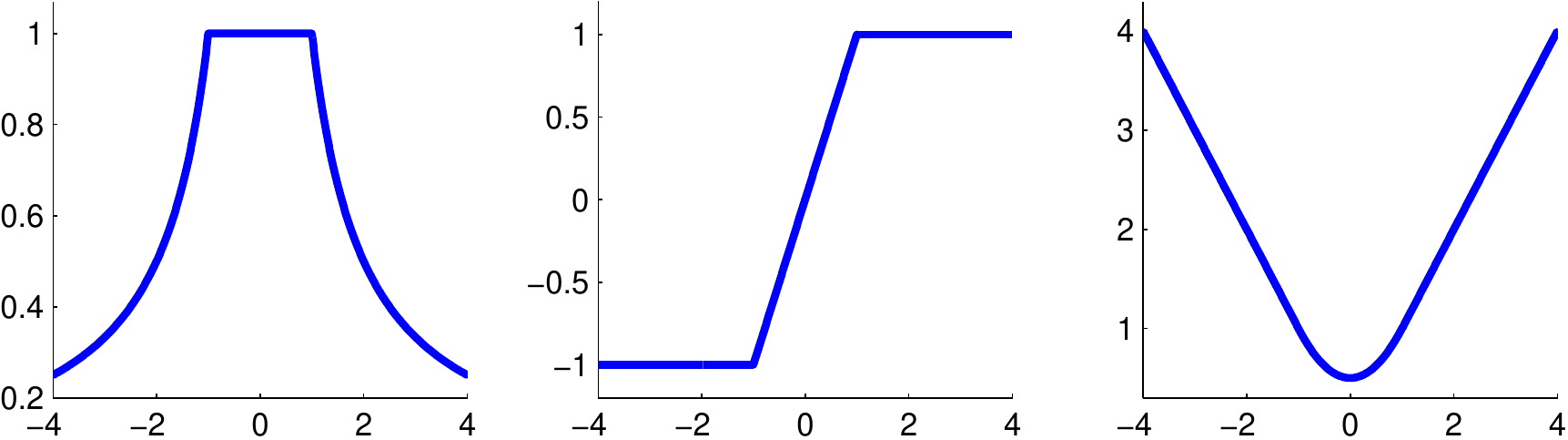}}
\subfigure[Tukey function]
{\label{tukey}\includegraphics[scale=0.53]{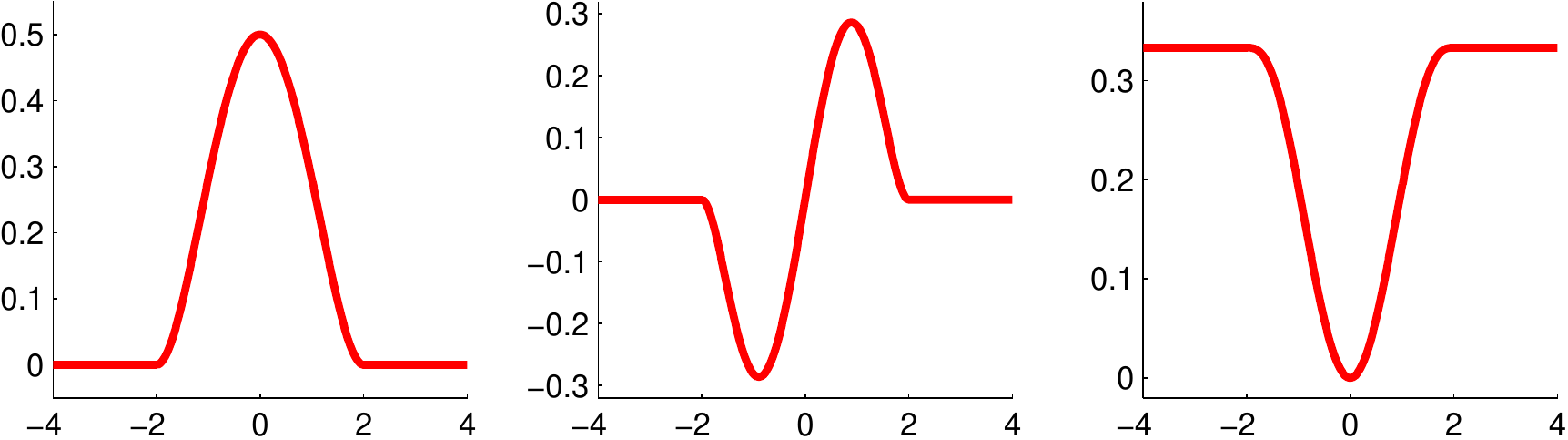}}
\caption[Comparing the Huber switching function with the Tukey function]
{Comparing the Huber switching function with the Tukey function: 
graphs of the edge-stopping function
$g(\sigma)$ are in the left column; graphs of
$\phi(\sigma)=\sigma g(\sigma)$ are in the middle column; graphs of
$\rho(\sigma)=\int \sigma g(\sigma) d\sigma$ are in the
right column.} \label{RegFun}
\end{figure}

Besides employing the implicit method to improve the quality of
pre-denoised images, we may wish to sharpen the
reconstructed image in a way that TV cannot provide \cite{sapiro}.
One possibility is to sharpen them by
making $R(m)$ in \eqref{fesd} gradually more and more non-convex
\cite{pema,Black,sapiro,Dur} and so reducing penalty on large jumps.
This can be done by replacing the Huber switching function depicted in Fig.~\ref{Rhuber}
with the Tukey function depicted in Fig.~\ref{tukey}.

The Tukey function, scaled similarly to the Huber function, is defined by
\begin{eqnarray}
\rho (\sigma) = \begin{cases}
\frac{1}{3} & |\sigma| \geq \hat \gamma , \cr (\frac{\sigma^2}
{\hat \gamma^2}-\frac{\sigma^4}{\hat \gamma^4}+\frac{\sigma^6}{3\hat \gamma^6}) & |\sigma| < \hat \gamma .
\end{cases}
\label{Turho}
\end{eqnarray}
Since
\begin{eqnarray*}
\rho'(\sigma)= \phi (\sigma) = \begin{cases}
0 & |\sigma| \geq \hat \gamma , \cr \frac{2\sigma}{\hat \gamma ^2}(1-(\frac{\sigma}{\hat \gamma})^2)^2 & |\sigma| < \hat \gamma ,
\end{cases}
\end{eqnarray*}
the resulting edge-stopping function is
\begin{eqnarray*}
g (\sigma) = \frac{\phi (\sigma)}{\sigma} = \begin{cases}
0 & |\sigma| \geq \hat \gamma , \cr \frac{2}{\hat \gamma ^2}(1-(\frac{\sigma}{\hat \gamma})^2)^2 & |\sigma| < \hat \gamma .
\end{cases}
\end{eqnarray*}
To ensure that the Tukey and Huber functions start rejecting outliers at the same value
we set 
$$ \hat \gamma = \sqrt{5}\gamma, $$
where $\gamma$ for the Huber function is still defined by \eqref{gammadef}.

\begin{figure}[tp]
\centering \subfigure[True image]
{\label{cam}\includegraphics[scale=0.35]{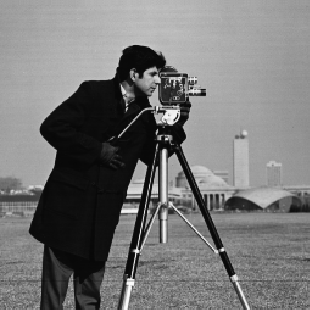}}
\subfigure[20\% noise]
{\label{cam20}\includegraphics[scale=0.35]{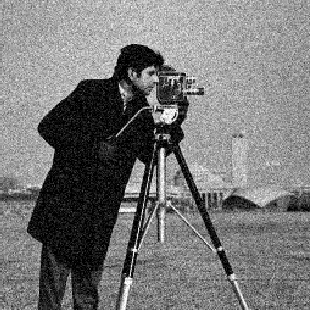}}
\centering \subfigure[Misfit = 18.93]
{\label{camSDe-4}\includegraphics[scale=0.35]{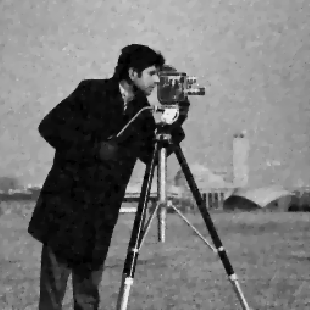}}
\subfigure[Misfit = 22.67]
{\label{camim}\includegraphics[scale=0.35]{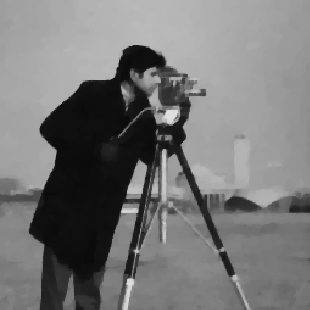}}
\subfigure[Misfit = 19.66]
{\label{camtu}\includegraphics[scale=0.35]{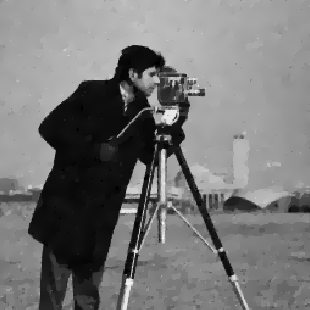}}
\caption{Smoothing and sharpening the $256 \times 256$ {\tt Cameraman} image: (a) true
image; (b) image corrupted by $20\%$ Guassian white noise serves as
data; (c) pre-denoised using 21 SD steps for a rougher relative
error tolerance of $10^{-4}$; (d) then denoised by 3 IRLS iterations
with $\beta = 0.15$ determined by \eqref{eibeta}, the total CPU time
= 3.4 + 5.2 = 8.6 sec; (e) alternatively,  sharpened using 10 SD steps with the Tukey
function starting from the result of (c), the total CPU time = 3.4 + 2.2 = 5.6 sec.}
\label{camhy}
\end{figure}

Thus, the influence of the Tukey function decreases all the way to zero. 
Comparing the two functions $\phi(\sigma)=\sigma g(\sigma)$ in the
horizontal center of Fig.~\ref{RegFun}, the Huber function gives all outliers a
constant weight of one whereas the Tukey function gives zero weight
to outliers whose magnitude is above a certain value. From such
shapes of $\phi$ we can correctly predict that smoothing with Tukey
produces sharper boundaries than smoothing with Huber. We can also
see how the choice of edge-stopping function acts to hold off excessive
smoothing: given a piecewise constant image where all
discontinuities are above a certain threshold, Tukey will leave the image
unchanged whereas Huber will not. Results in Figs.~\ref{lenatu} and
\ref{camtu} confirm our predictions. After a quick pre-denoising
with Huber, only 10 SD steps with Tukey result in obviously sharper
discontinuities, i.e., image edges, than those without switching the regularization operator; 
see Figs.~\ref{lenahy} and \ref{camhy}.

It is important to note that a rough
pre-denoising is {\em necessary} here. It helps us avoid strengthening
undesirable effects of heavy noise by using the Tukey function.

From an experimental point of view, we recommend to apply the explicit-implicit LSD
scheme if a smooth image is desired;
otherwise, employ the explicit Tukey regularization, starting from the result
of the rough explicit Huber regularization, to get a sharper version.

\section{Deblurring}
\label{sec:deblurring}

Here we consider the deblurring problem discussed in \cite{vogelbook,chsh,Hardy}.
The blurring of an image can be caused by many factors: (i) movement
during the image capture process, by the camera or, when long
exposure times are used, by the subject; (ii) out-of-focus optics,
use of a wide-angle lens, atmospheric turbulence, or a short
exposure time, which reduces the number of photons captured; (iii)
scattered light distortion in confocal microscopy. Mathematically,
in most cases blurring can be linearly modeled to be {\em shift-invariant} with a
point spread function (PSF), denoted by $f(x,y)$.
Further, it is well known in signal processing and systems theory \cite{OppSch,OppWil} 
that a shift-invariant
linear operator must be in the form of convolution, written as
\begin{eqnarray}
\label{convo}
F(m) = f(x,y) * m(x,y) = \int_{\R^2}f(x-x',y-y')m(x',y')dx'dy'.
\end{eqnarray}
So, an observed blurred image $b$ is related to the ideal sharp image $m(x,y)$ by
$$b = f(x,y) * m(x,y) +\epsilon ,$$
where $*$ denotes convolution product and the point spread function $f(x,y)$ may vary in space.
Thus, the matrix-vector multiplication
$Jm$ represents a discrete model of the distortion operator
\eqref{convo} convolved by the PSF.


In the spatial domain, the PSF describes the degree to which an optical system
blurs or spreads a point of light. The PSF is the inverse Fourier
transform of the optical transfer function (OTF). In the frequency domain, the
OTF describes the response of a linear, position-invariant
system to an impulse. 
The distortion is created by
convolving the PSF with the original true image, see
\cite{vogelbook,chsh}. Note
that distortion caused by a PSF is just one type of data degradation,
and the clear image $m_{true}$ generally does not exist in
reality. This image represents the result of
perfect image acquisition conditions. 
Nonetheless, in our numerical experiments we have a ``ground truth''
model $m_{true}$ which is used to synthesize data and judge the quality of reconstructions.

In the implementation we apply the same discretization as in Chapter 5 of
\cite{vogelbook}, and then the convolution \eqref{convo} is discretized into a
matrix-vector multiplication, yielding a problem of the form \eqref{1}, \eqref{Jm},
where $J$ is an $N\times N$ symmetric,
doubly block Toeplitz matrix. Such a
blurring matrix $J$ can be constructed by a Kronecker product.
However, $J$ is now a full, large matrix, and avoiding its explicit construction
and storage is therefore desirable.
Using a gradient descent algorithm
we actually only need two matrix-vector products to form $J^T(Jm)$, and there is no reason
to construct or store the matrix $J$ itself. Moreover, it is possible to use a
2D fast Fourier transform (FFT) algorithm to reduce the
computational cost of the relevant matrix-vector multiplication from
$\Om (N^2)$ to $\Om (N\log N)$.
Specifically, after discretizing the integral operator \eqref{convo} in the form
of convolution, we have a fully discrete model
$$b_{i,j}=\sum^{n}_{p=0}\sum^{n}_{q=0}f_{i-p,j-q}m_{p,q} + \epsilon_{i,j},\quad 0\leq i,\; j\leq n ,$$
where $\epsilon_{i,j}$ denotes random noise at the grid location
$(i,j)$. In general, the discrete PSF $\{f_{i,j}\}_{i,j=0}^n$ is
2D-periodic, defined as
\begin{eqnarray*}
f_{i,j} &=& f(i',j), \quad\mbox{whenever}\quad i = i' ~\texttt{mod}~ n+1,\\
f_{i,j} &=& f(i,j'),\quad\mbox{whenever}\quad j = j' ~\texttt{mod}~ n+1.
\end{eqnarray*}
This suggests we only need consider $f\in \R^{(n+1)\times (n+1)}$,  because by periodic extension we can easily
get $(n+1,n+1)$-periodic arrays $f^{ext}$, for which
$$f^{ext}_{i,j}=f_{i,j}, \quad\mbox{whenever}\quad 0\leq i,\; j\leq n .$$
Then the FFT algorithm yields
\begin{eqnarray*}
f^{ext} * m = (n+1)\,\mathcal{F}^{-1}\{\mathcal{F}(f)\,.*\,\mathcal{F}(m)\},
\end{eqnarray*}
where $\mathcal{F}$ is the discrete Fourier transform and
$.*$ denotes component-wise multiplication. 

Thus, we consider the gradient descent algorithm \eqref{gradesc}, starting from the data $m^0 = b$,
and compare the step size choices
\eqref{sd} vs. \eqref{lsd}.
As before, strictly speaking, 
these would be steepest descent and lagged steepest descent only if $L$ defined in \eqref{Lm}
were constant, i.e., using least-squares regularization. 
But we proceed to freeze $L$ for this purpose anyway, which amounts to a lagged diffusivity approach.

One difference from the denoising problem is that here we do not have an easy
tool for determining the regularization parameter $\beta$, and it is determined
experimentally instead. However, for the problems discussed below this turns out not 
to be a daunting task.

\begin{figure}[tp]
\centering
\subfigure[True image]
{\label{boat}\includegraphics[scale=0.4]{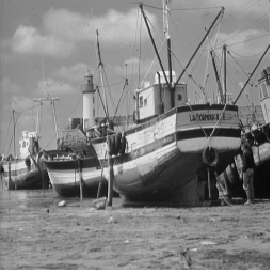}}\qquad
\subfigure[{\sc motion} blur]
{\label{boatmotion}\includegraphics[scale=0.4]{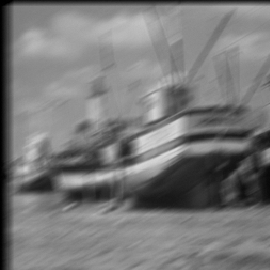}}\\
\subfigure[$\beta = 10^{-3}$, 23 LSD iters]
{\label{boatbetae-3}\includegraphics[scale=0.4]{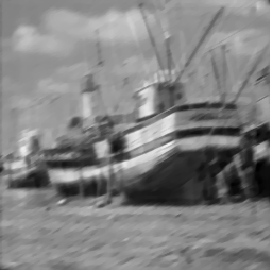}}\quad
\subfigure[$\beta = 10^{-4}$, 33 LSD iters]
{\label{boatbetae-4}\includegraphics[scale=0.4]{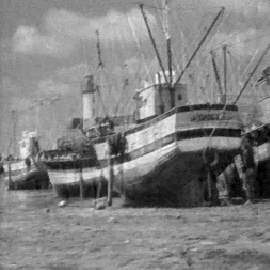}}\quad
\subfigure[$\beta = 10^{-5}$, 38 LSD iters]
{\label{boatbetae-5}\includegraphics[scale=0.4]{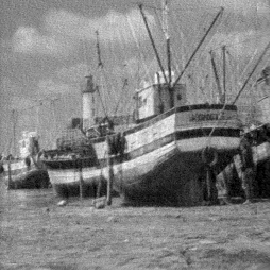}}
\caption[{\sc type = `motion', len = 15, theta = 30}, $\eta = 1$]{{\sc type = `motion', len = 15, theta = 30}, $\eta = 1$.}
\label{boat_motion}
\end{figure}

\begin{figure}[tp]
\centering
\begin{tabular}{c}
\\ \\ \\ \\
\includegraphics[scale=0.58]{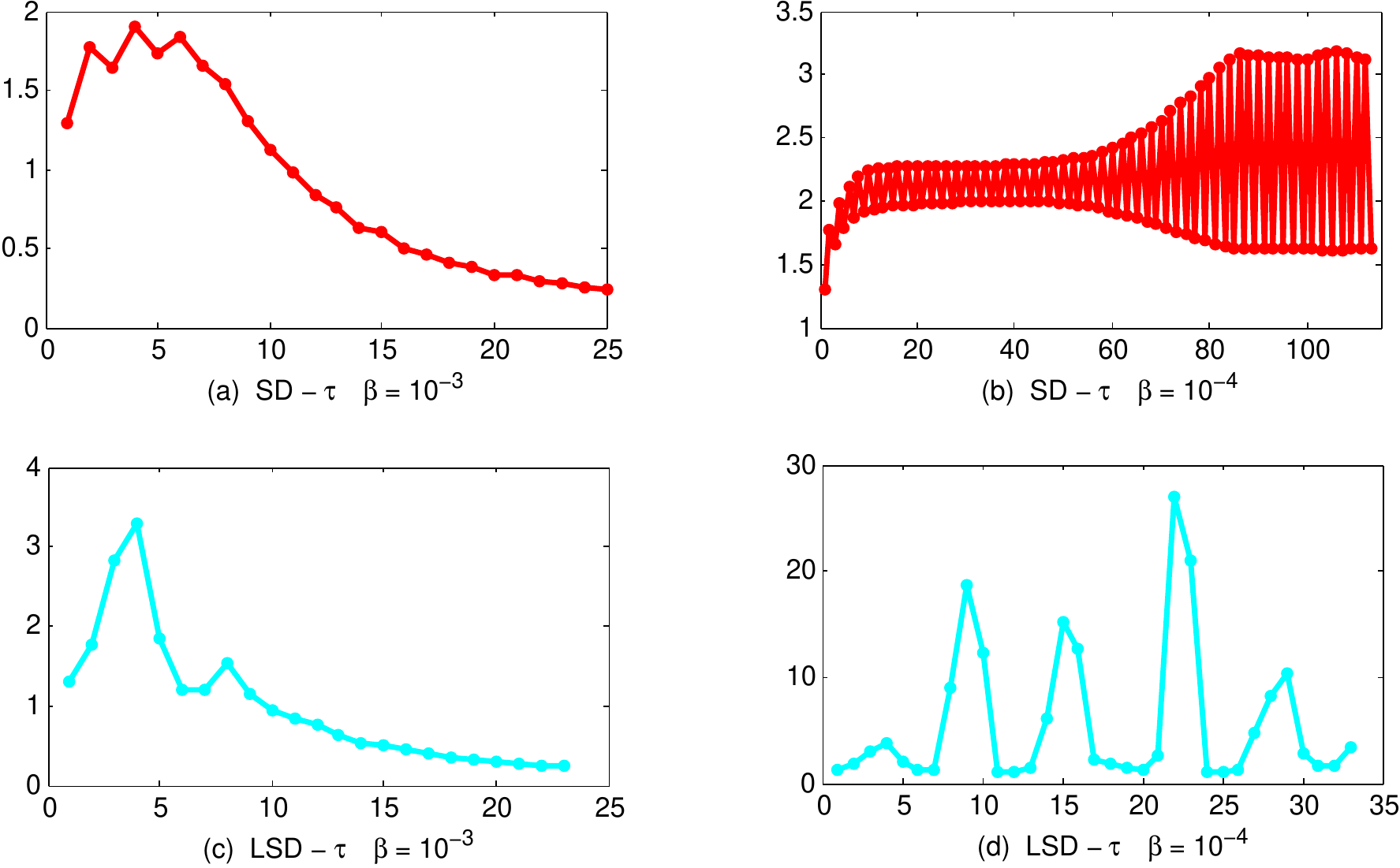}
\end{tabular}
\caption[SD step sizes vs. LSD step sizes for the deblurring example]{SD step size \eqref{sd} vs. LSD step size \eqref{lsd} 
for the deblurring example in Fig.~\ref{boat_motion} under the relative error tolerance of $10^{-4}$: 
(a) and (c) compare SD with LSD for $\beta = 10^{-3}$, 25 SD iters vs. 23 LSD iters; 
(b) and (d) compare them for $\beta = 10^{-4}$, 113 SD iters vs. 33 LSD iters.
Note the horizontal scale (number of iterations) difference between (b) and (d).}
\label{boatSDLSD}
\end{figure}

\begin{figure}[tp]
\centering
\subfigure[True image]
{\label{house}\includegraphics[scale=0.4]{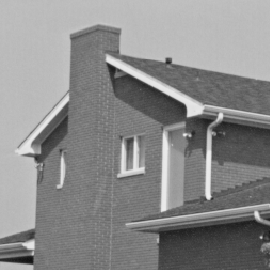}}\quad
\subfigure[{\sc log} blur]
{\label{houselog}\includegraphics[scale=0.4]{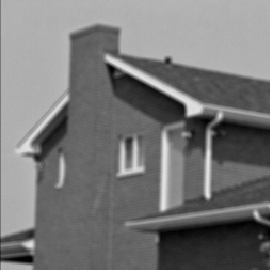}}\quad
\subfigure[114 SD or 31 LSD iters]
{\label{Dhouse}\includegraphics[scale=0.4]{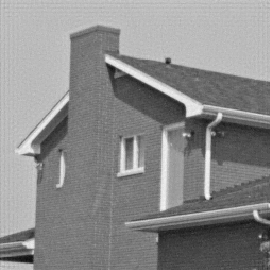}}
\caption[{\sc type = `log', sigma = 0.5}, $\eta = 1$, $\beta = 10^{-4}$]{{\sc type = `log',  hsize = [512, 512], sigma = 0.5}, $\eta = 1$, $\beta = 10^{-4}$.}
\label{house_log}
\end{figure}

\begin{figure}[tp]
\centering
\subfigure[True image]
{\label{hill}\includegraphics[scale=0.4]{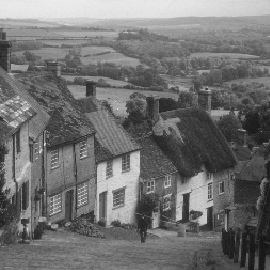}}\quad
\subfigure[{\sc disk} blur]
{\label{hilldisk}\includegraphics[scale=0.4]{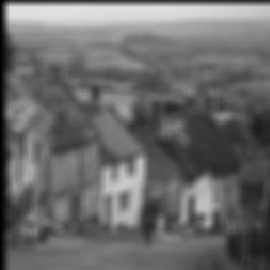}}\quad
\subfigure[99 SD or 26 LSD iters]
{\label{Dhill}\includegraphics[scale=0.4]{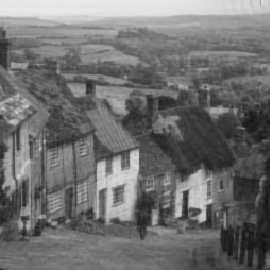}}
\caption[{\sc type = `disk',  radius = 5}, $\eta = 1$, $\beta = 10^{-4}$]{{\sc type = `disk',  radius = 5}, $\eta = 1$, $\beta = 10^{-4}$.}
\label{hill_disk}
\end{figure}

\begin{figure}[tp]
\centering
\subfigure[True image]
{\label{sate}\includegraphics[scale=0.4]{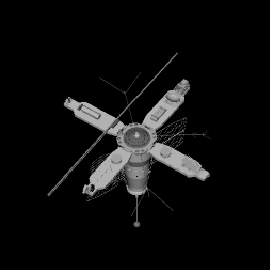}}\quad
\subfigure[{\sc unsharp} blur]
{\label{sateunaharp}\includegraphics[scale=0.4]{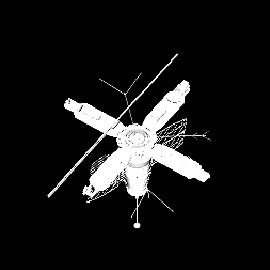}}\quad
\subfigure[17 SD or 15 LSD iters]
{\label{Dsate}\includegraphics[scale=0.4]{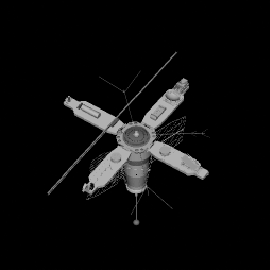}}
\caption[{\sc type = `unsharp',  alpha = 0.2}, $\eta = 1$, $\beta = 10^{-4}$]{{\sc type = `unsharp',  alpha = 0.2}, $\eta = 1$, $\beta = 10^{-4}$.}
\label{sate_unsharp}
\end{figure}

\begin{figure}[tp]
\centering \subfigure[True image]
{\label{toy}\includegraphics[scale=0.39]{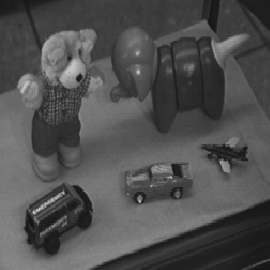}}
\subfigure[{\sc gaussian} blur]
{\label{toy5}\includegraphics[scale=0.39]{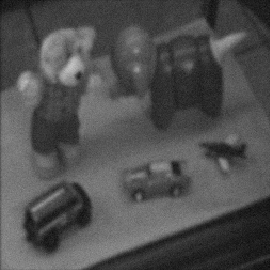}}
\subfigure[Noise amplification]
{\label{Dtoy}\includegraphics[scale=0.39]{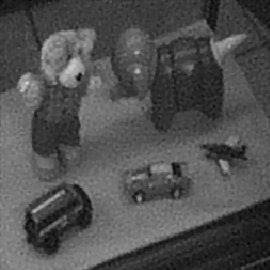}}
\subfigure[Pre-denoised]
{\label{toypre}\includegraphics[scale=0.39]{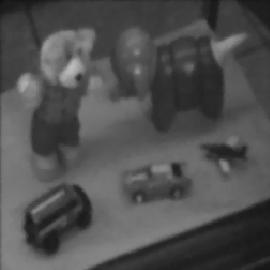}}
\subfigure[Deblurred]
{\label{toyblur}\includegraphics[scale=0.39]{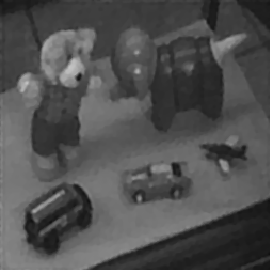}}
\subfigure[Sharpened]
{\label{toysharp}\includegraphics[scale=0.39]{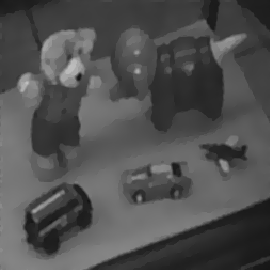}}
\caption[{\sc type = `gaussian', sigma = 1.5},
$\eta = 5$, $\beta = 5\times 10^{-4}$]{{\sc type = `gaussian', hsize = [512, 512], sigma = 1.5}:
(b) data $b$ corresponding to $\eta = 5$; (c) directly deblurring using $\beta = 5\times 10^{-4}$ fails;
(d) pre-denoising: for the relative error tolerance of
$10^{-4}$, 9 LSD steps of \eqref{fesd} are required;
(e) 22 steps of \eqref{gradesc} with $\beta = 5\times 10^{-4}$ and LSD
\eqref{lsd} are required for the next deblurring; (f) 10
sharpening steps with the Tukey function \eqref{Turho} follow. The total CPU time
= 16.1 sec.} \label{toy_gau}
\end{figure}

\begin{figure}[tp]
\centering \subfigure[True image]
{\label{pepper}\includegraphics[scale=0.39]{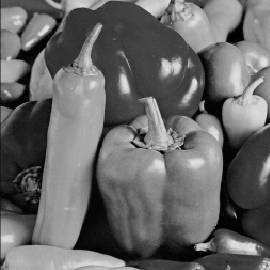}}
\subfigure[{\sc laplacian} blur]
{\label{pepper10}\includegraphics[scale=0.39]{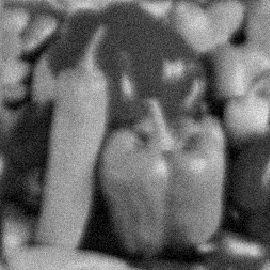}}
\subfigure[Noise amplification]
{\label{Dpepper}\includegraphics[scale=0.39]{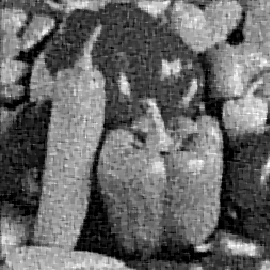}}
\subfigure[Pre-denoised]
{\label{pepperpre}\includegraphics[scale=0.39]{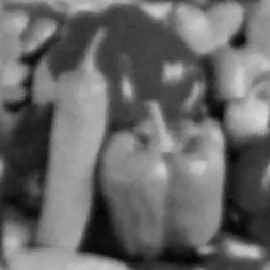}}
\subfigure[Deblurred]
{\label{pepperblur}\includegraphics[scale=0.39]{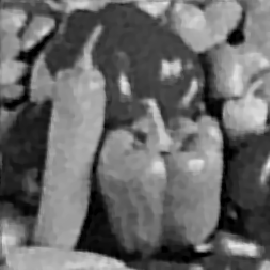}}
\subfigure[Sharpened]
{\label{peppersharp}\includegraphics[scale=0.39]{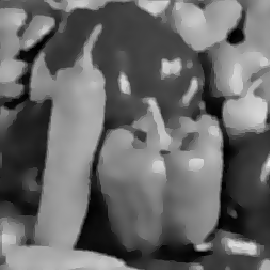}}
\caption[{\sc type = `laplacian', alpha = 0.2}, $\eta = 10$, $\beta
= 10^{-3}$]{{\sc type = `Laplacian', alpha = 0.2}:
(b) data $b$ corresponding to $\eta = 10$; (c) directly deblurring using $\beta = 10^{-3}$ fails;
(d) pre-denoising: for the relative error tolerance of
$10^{-4}$, 10 LSD steps of \eqref{fesd} are required;
(e) 25 steps of \eqref{gradesc} with $\beta = 10^{-3}$ and LSD
\eqref{lsd} are required for the next deblurring; (f) 10
sharpening steps with the Tukey function \eqref{Turho} follow. The total CPU time
= 14.5 sec.}
\label{pepper_lap}
\end{figure}

To illustrate and analyze our deblurring algorithm, we generate some degraded data at first.
We use the {\sc Matlab} function {\sc fspecial} to create a variety of correlation
kernels, i.e.,~PSFs, and then deliberately blur clear images by convolving them with these different PSFs. The function
{\sc fspecial(type,parameters)} accepts a filter type plus additional modifying parameters particular to the type of filter chosen.
Thus, {\sc fspecial(`motion',len,theta)} returns a filter to approximate, once convolved with an image, the linear motion of a camera by
{\sc len} pixels with an angle of {\sc theta} degrees in a counterclockwise direction, which therefore becomes a vector for horizontal and vertical motions (see Fig.~\ref{boat_motion}); {\sc fspecial(`log',hsize,sigma)} returns a rotationally symmetric Laplacian of Gaussian filter of size
{\sc hsize} with standard deviation {\sc sigma} (see Fig.~\ref{house_log});
{\sc fspecial(`disk',radius)} returns a circular mean filter within the square matrix of side 2\,{\sc radius}+1 (see Fig.~\ref{hill_disk}); {\sc fspecial(`unsharp',alpha)} returns a $3\times 3$ unsharp contrast enhancement filter, which enhances edges and other high frequency components by subtracting a smoothed unsharp version of an image from the original image, and the shape of which is controlled by the parameter
{\sc alpha} (see Fig.~\ref{sate_unsharp}); {\sc fspecial(`gaussian',hsize,sigma)} returns a rotationally symmetric Gaussian low-pass filter of size
{\sc hsize} with standard deviation {\sc sigma} (see Fig.~\ref{toy_gau}); {\sc fspecial(`laplacian',alpha)} returns a $3 \times 3$ filter approximating the shape of the two-dimensional Laplacian operator and the parameter {\sc alpha} controls the shape of the Laplacian (see Fig.~\ref{pepper_lap}).

All six images presented and used in this section are $256 \times 256$.
For the first four experiments, we only add a small amount of
random noise into blurred images, say 1\% ($\eta =1$), and stop
deblurring when the relative error norm \eqref{relaerr} is below
$10^{-4}$. For a given PSF, the only parameter required
is the regularization parameter $\beta$. From Fig.~\ref{boat_motion}
we can clearly see that the smaller $\beta$ is, the sharper the
restored solution is, including both image and noise. So the
{\tt Boat} image reconstructed with $\beta = 10^{-3}$ in Fig.~\ref{boatbetae-3}
still looks blurry and that cannot be improved by running more
iterations. The {\tt Boat} image deblurred with $\beta = 10^{-5}$ in
Fig.~\ref{boatbetae-5} becomes much clearer; however, unfortunately,
such a small $\beta$ also brings the undesirable effect of noise
amplification. The setting $\beta=10^{-4}$ seems to generate the
best approximation of the original scene, and so it does in the 
following three deblurring experiments. 

As in the case of denoising, the LSD step size selection \eqref{lsd} usually yields faster convergence than SD. 
When $\beta$ is small, e.g., $10^{-4}$ or less, 
the step size \eqref{sd} is very close to the strict steepest descent selection obtained for a constant $L$,
and so the famous two-periodic cycle of \cite{akaike} (see also \cite{asdohusv}) appears in the step sequence, 
resulting in a rather slow convergence; see Fig.~\ref{boatSDLSD}(b). 
The lagged step size \eqref{lsd} breaks this cycling pattern (see Fig.~\ref{boatSDLSD}(d)), 
providing a much faster convergence for the same error tolerance. 
Moreover, with the same parameter $\beta$, the reconstructed images using both SD and LSD are quite comparable, 
and it is difficult to tell any differences between them by the naked eye. 
For $\beta=10^{-5}$, 335 SD steps are required to reach a result comparable to Fig.~\ref{boatbetae-5} 
and the corresponding CPU time is 441.6 sec. 
Using LSD we only need 48.7 sec, reflecting the fact that 
the CPU time is roughly proportional to the number of steps required.

Figs.~\ref{house_log} -- \ref{sate_unsharp} reinforce our previous observations that the
gradient descent deblurring algorithm with LSD step selection
\eqref{lsd} works very well, and that the improvement of LSD over SD
is even more significant here than in the case of denoising.
This is especially pronounced when a small value of $\beta$ must be chosen 
and when accuracy considerations require
more than 20 or so steepest descent steps.

In \cite{asdohusv} we have discussed other faster gradient descent methods.
The half-lagged steepest descent (HLSD) method \cite{rasv,fmmr} 
was generally found there to consistently be at par with LSD, 
while other variants performed somewhat worse.
In the present article we have applied HLSD in place of LSD for the above four examples,
where the step size selection counts most.
With HLSD the formula \eqref{sd} is applied only at
each even-numbered step and then the same step size 
gets reused in the following odd-numbered one.
The results were found to be again comparable to those using LSD,
both in terms of efficiency and in terms of quality.

We have also experimented with a version of CG where we locally pretend,
as in \eqref{steps}, that the minimization problem is quadratic.
However, the CG method is well-known to be more sensitive than gradient
descent to violations of its premises, and its iteration counts when
applied to each of the examples in Figs.~\ref{boat_motion} -- \ref{hill_disk}
were consistently about 20\% higher than the better of LSD and HLSD.
\vspace{0.7cm}

\noindent{\bf Deblurring noisier images}
\vspace{0.3cm}

The operation of deblurring essentially sharpens the image, whereas denoising
essentially smooths it. Thus, trouble awaits any algorithm when both significant 
blur and significant noise are present in the given data set.

In our present setting, if we add more noise to the blurred images when synthesizing
the data, say $\eta \geq 5$, then
directly running the deblurring gradient descent algorithm as above may fail due
to the more severe effect of noise amplification; see,
e.g.,~Figs.~\ref{Dtoy} and \ref{Dpepper}. 
After a few iterations, the
restored image can have a speckled appearance, especially for a
smooth object observed at low signal-to-noise ratios. These speckles
do not represent any real texture in the image, but are artifacts
of fitting the noise in the given image too closely. Noise amplification can be
reduced by increasing the value of $\beta$, but this may result in a
still blurry image, as in Fig.~\ref{boatbetae-3}. Since the
effects of noise and blur are opposite, a quick and to some extent effective
remedy is {\em splitting}, described next.

At first, we only employ denoising up to a coarser tolerance,
e.g.,~$10^{-4}$. This can be carried out in just a few steps of \eqref{fesd}
with either the SD or LSD step size selection. Starting with the lightly
denoised image, as in Figs.~\ref{toypre} and \ref{pepperpre}, we next apply the
deblurring algorithm \eqref{gradesc} with the LSD step sizes to correct PSF
distortion. Since now the noise level becomes higher, we slightly increase 
the value of the regularization parameter and apply $\beta = 5\times 10^{-4}$ for the {\tt Toy} 
example and $\beta = 10^{-3}$ for the {\tt Pepper} example.
Observe that, even though some speckles still appear on the
image cartoon components, the results presented in
Figs.~\ref{toyblur} and \ref{pepperblur} are much more acceptable
than those in Figs.~\ref{Dtoy} and \ref{Dpepper}, which were deblurred by the
same number of iterations without pre-denoising. Finally, we can use
the Tukey regularization in \eqref{fesd} 
to further improve the reconstruction,
carefully yet rapidly removing unsuitable speckles and
enhancing the contrast, i.e., {\em sharpening}. The results are demonstrated in Figs.~\ref{toysharp} and
\ref{peppersharp}. 
The total CPU time, given in the captions of
Figs.~\ref{toy_gau} and \ref{pepper_lap}, clearly shows the
efficiency of this hybrid deblurring-denoising scheme.

\section{Conclusions}
\label{sec:conclusion}

In this paper we have examined the effect of replacing steepest descent (SD)
by a faster gradient descent algorithm, specifically, lagged steepest descent (LSD),
in the practical context of image deblurring and denoising tasks.
We have also proposed several highly efficient schemes for carrying
out these tasks, independently of the step size selection.

Our general conclusion is that in situations where many (say, over 20)
steepest descent steps are
required, thus building slowness into the solution procedure, the faster gradient descent method
offers substantial advantages.

Specifically, four scenarios have been considered.
The first is a straightforward denoising process using anisotropic diffusion \cite{asdohusv}.
Here the LSD step selection offers an efficiency improvement by a factor of roughly $3$.

In contrast, the second denoising scenario does not allow slowness buildup by SD because
after a quick rough denoising we switch to an implicit method, with a good estimate for $\beta$
at hand, or to a sharpening phase using \eqref{Turho}. The resulting method is new and effective,
although not because of a dose of LSD.

Switching to the more interesting and challenging deblurring problem, 
the third scenario envisions the presence of little additional noise,
so we directly employ the gradient descent method \eqref{gradesc} with $J$ as described
in Section~\ref{sec:deblurring}, $R$ given by \eqref{hubereg} and \eqref{gammadef},
and $\beta = 10^{-4}$.
This allows for slowness buildup when using SD step sizes, and the faster LSD variant
then excels, becoming up to 10 times more efficient.
The HLSD variant is overall as effective as LSD.

Finally, in the presence of significant noise effective deblurring becomes a harder task.
We propose a splitting approach whereby we switch between the previously
developed denoising and deblurring algorithms.
This again creates a situation where LSD does not contribute much improvement over SD.
The splitting approach has been demonstrated to be relatively effective, although none
of the reconstructions in Fig.~\ref{pepper_lap}, for instance, is amazingly good.
The problem itself can become very hard to solve satisfactorily, 
unless some rather specific knowledge about the
noise is available and can be used to carefully remove it before deblurring begins.

A future problem to be considered concerns the case where the PSF causing blurring
is not known.   
  
\bibliographystyle{spmpsci}

\bibliography{biblio}
\end{document}